\documentclass[runningheads]{llncs}

\usepackage{amsmath}
\usepackage{amssymb}
\usepackage{algorithm}
\usepackage{algpseudocode}
\usepackage{blindtext}
\usepackage{comment}
\usepackage[all]{xy}
      
\usepackage{listings}
\usepackage{color}

\newcommand{\C}[1]{\mbox{\lstinline`#1`}}

\definecolor{dkblue}{rgb}{0,0.1,0.5} 
\definecolor{lightblue}{rgb}{0,0.5,0.5} 
\definecolor{dkgreen}{rgb}{0,0.4,0} 
\definecolor{dk2green}{rgb}{0.4,0,0} 
\definecolor{dkviolet}{rgb}{0.6,0,0.8}
\definecolor{shadethmcolor}{rgb}{0.9, 0.9,1}

\usepackage{multirow}
\usepackage{xcolor}

\usepackage{graphicx}
\usepackage{todonotes}

\begin{document}

\title{Neural Network Verification for the Masses\\ (of AI graduates)\thanks{Supported by EPSRC DTA Scholarship and EPSRC Platform Grant EP/N014758/1.}}

\author{Ekaterina Komendantskaya\inst{1}
  \and
  Rob Stewart\inst{1}
  \and
  Kirsy Duncan\inst{1}
  \and
  Daniel Kienitz\inst{1}
  \and
  Pierre Le Hen\inst{1}
  \and
  Pascal Bacchus\inst{1}
}

\authorrunning{E. Komendantskaya et al.}

\institute{Mathematical and Computer Sciences  \\ Heriot-Watt University \\ Edinburgh, UK}

%
\maketitle
\begin{abstract}
 
  Rapid development of AI applications has stimulated demand for, and has given rise to, the
  rapidly growing number and diversity of AI MSc degrees. AI and Robotics
  research  communities, industries and students are becoming increasingly aware of the
  problems caused by unsafe or insecure AI applications. Among them, perhaps the most famous
  example is vulnerability of deep neural networks to ``adversarial attacks''.
  Owing to wide-spread use of neural networks in all areas of AI, this problem is seen as
  particularly acute and pervasive.

  Despite of the growing number of research papers about safety and security vulnerabilities of AI
  applications, there is a noticeable shortage of accessible tools, methods and
  teaching materials for incorporating verification into AI programs. 
  
  LAIV -- the Lab for AI and Verification --  is a newly opened research lab at
  Heriot-Watt university that engages AI and Robotics MSc students in verification
  projects, as part of their MSc dissertation work. In this paper, we will report on
  successes and unexpected difficulties LAIV faces, many of which arise from limitations of existing programming languages used for verification.
  We will discuss future directions
  for incorporating verification into AI degrees.

\keywords{AI MSc degrees  \and Neural Network Verification \and AI Verification \and Python \and SMT Solvers \and Interactive Theorem Provers}
\end{abstract}

\section{Introduction}

AI applications have become pervasive: from mobile phones and home appliances to autonomous cars and stock markets --
we are served by a range of intelligent algorithms, of which one prominent group is represented under an umbrella term \emph{neural networks}.
In the recent past, neural networks have been shown to match a human-level performance in specific domains such as
speech recognition and natural language processing \cite{Hinton2012}, image classification \cite{KrizhevskySH17}
and reinforcement learning \cite{abs-1807-01281}, winning their prominent place among tools used in both research and industry. 

It comes at little surprise then, that increasing number of undergraduate and postgraduate students choose AI and machine learning
as their specialism. This global trend is reflected in the university degrees. 
In partnership with Edinburgh University, Heriot-Watt university is a home for National \emph{Robotarium}\footnote{\url{https://www.edinburgh-robotics.org/robotarium}}, the Doctoral Training
Center \emph{``Edinburgh Center for Robotics''}\footnote{\url{https://www.edinburgh-robotics.org/}}, Robotics Lab\footnote{\url{http://www.macs.hw.ac.uk/RoboticsLab/}} and a number of large research projects.
Over a dozen of AI and Robotics MSc degrees have been opened
at Heriot-Watt, with specialisms ranging from general AI and Robotics degrees to specialised degrees such as ``Robot Human Interaction'' and ``Speech with Multimodal Interaction''.
Students undertake courses covering both symbolic and statistical AI, with topics ranging from
AI planning languages to Bayesian learning and deep neural networks.

With the growing number of AI applications, there is also a growing concern that only a small
proportion of them are \emph{verified} to be trustworthy, safe and secure. This concern stimulated
recent research in AI verification generally and 
neural network verification in particular.
Research in neural network verification splits into two main groups of research questions: 
\begin{itemize}
\item  \emph{Concerning properties of learning algorithms in general}: e.g. how well does a given learning algorithm perform?
  do trained  neural networks generalise well to classify yet unseen data? The answers include proofs of properties of generalisation bounds,
  equality of neural networks, properties of neural network architectures.  
  Good examples are~\cite{BS19,BMF18}.  Usually, these verification projects are conducted in interactive theorem provers (ITPs), such as Coq~\cite{Coq},
  as they benefit from Coq's rich higher-order language and well-developed proof libraries.
  Unfortunately, Coq as a language does not offer machine learning support comparable to e.g. Python.
  Also, compared to automated provers,  proof automation is rather limited in ITP.

\item  \emph{Concerning specific neural network applications or their specific deployment}: e.g. given a trained neural network,
  is it robust to \emph{adversarial attacks}~\cite{HuangKWW17,SinghGPV19}?

 Adversarial attacks in this setting are usually given by adversarial examples, which are in turn given by
 non-random small-scale perturbations of an input data on which the neural network was trained.  These perturbations
 lead the neural network to output a wrong classification or regression result \cite{szegedy_2014_intriguing_properties_of_nns}.
\end{itemize}
 
 	\begin{figure}[t]
		\centering
		\includegraphics[width=.3\textwidth]{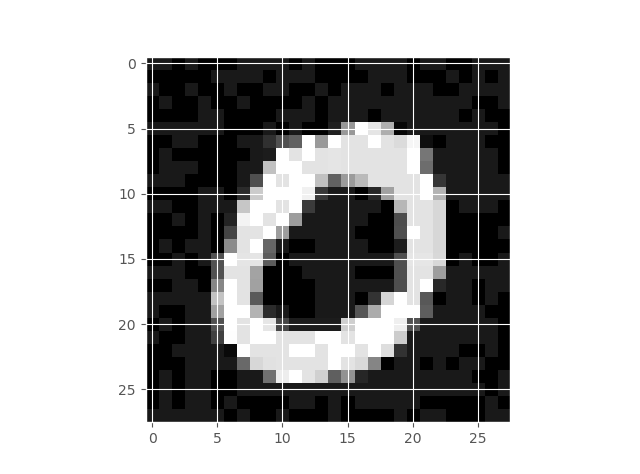}\hfill
		\includegraphics[width=.3\textwidth]{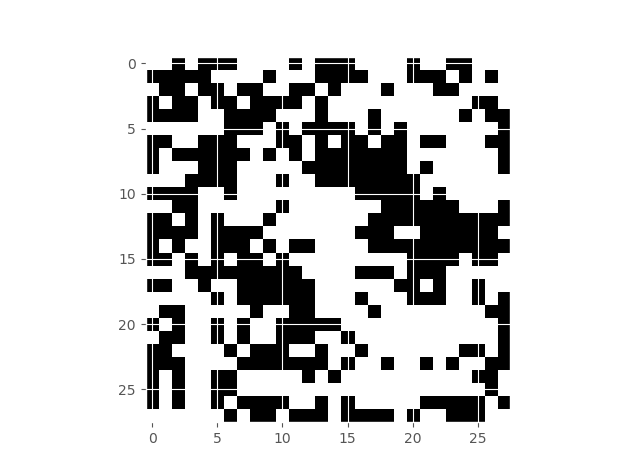}\hfill
		\includegraphics[width=.3\textwidth]{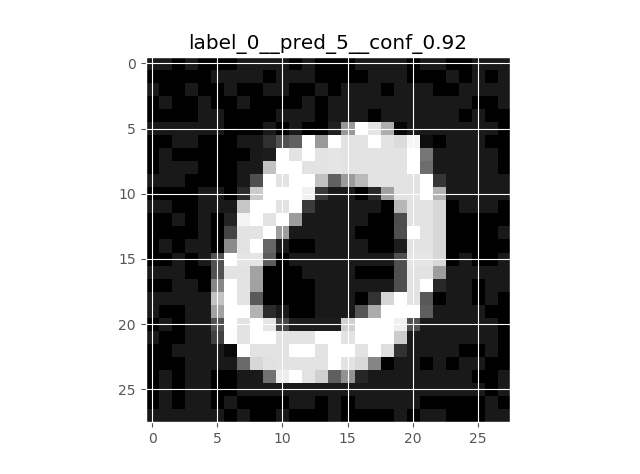}
		\caption{Given a trained neural network and a correctly classified image of ``0'' on the left,
                  we can create a perturbation $\eta$ (middle) to the original image so that the same neural network predicts a ``3" with 92\% confidence for the modified image (right).}
		\label{figure:adv_example}
	\end{figure}

	\begin{example}[Adversarial attack]\label{ex:adv}
          Figure \ref{figure:adv_example} shows how a seemingly insignificant  perturbation to the original
          image causes neural network to missclassify the image. It uses the MNIST data set~\cite{lecun_mnist}, a famous benchmark data set
          for hand-written digit recognition.
	\end{example}

         Adversarial attacks were successful
         at manipulating all kinds of neural networks, including recurrent~\cite{PapernotMSH16} and
         convolutional~\cite{abs-1812-03303} neural networks. In addition, adversarial examples were successfully crafted for all domains in
 which neural networks have been applied. See e.g.~\cite{SP18} for a comprehensive survey.

        The second kind of verification projects usually uses Python for the machine learning part and SMT solvers (such as Z3~\cite{MouraB08})
        for verification component, and often relies on Python's Z3 API. Thanks to this combination, these projects benefit from
        both comprehensive machine learning libraries of Python and the automation offered by Z3.
        Among disadvantages of this framework is fragility of code conversion between Python and Z3, and
        limitations on the generality of neural network properties that can be expressed in this setting.
  

        Despite of the growing volume of research and the growing industrial awareness,
        verification is not routinely taught as part of AI and Robotics BSc or  MSc degrees. 
        As a result, postgraduate and undergraduate students increasingly choose verification-related subjects
        for their dissertations. 
Thus, in the 2018-2019 MSc intake, we saw a surging interest from AI and Robotics MSc students in taking up
verification projects as part of their MSc dissertations. \emph{LAIV -- Lab for AI and Verification}\footnote{LAIV webpage: www.LAIV.uk}
-- was founded
in order to support AI, Robotics and Data Science MSc students venturing into verification field for their dissertations.
It  provides additional support in the forms of seminars and reading groups and
 thus compensates for absent verification courses in the MSc degrees.

In this paper, we reflect on the design space of programming languages and tools available
for AI and machine learning specialists and students who want to master neural network verification
in a \emph{lightweight}  manner, i.e. not as part of  their main specialism. 
Starting with some necessary background definitions in Section~\ref{sec:bg}, we proceed with introducing, by means of an easy running example,
three verification methods:

\begin{itemize}
\item  automated verification of neural networks via Python's Z3 API  (Section~\ref{sec:Py}),
\item interactive and functional  approach to neural network verification in Coq (Section~\ref{sec:Coq}), and
\item hybrid approach via $F^*$~\cite{MartinezADGHHNP19}, a novel functional language with capacity to delegate some
  theorem proving to Z3 (Section~\ref{sec:fstar}).  
  \end{itemize}
  We reflect on advantages and disadvantages of these approaches, both in terms of
  verification methodology and student experience.
  Finally, in Section~\ref{sec:concl} we conclude and outline some future directions.

\section{Background: Neural Networks and their Robustness}\label{sec:bg}

\begin{figure}[t]
		\centering
		\includegraphics[width=.15\textwidth]{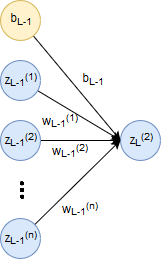} 
             \ \ \ \ \  \ \ \ \ 	\includegraphics[width=.35\textwidth]{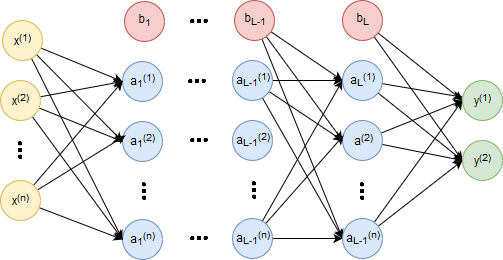} 
		\caption{\textbf{Left:} An example of a neuron in the last hidden layer $L$: Its potential is $z_L^2 = (\sum_{i=1}^{n} w^{(i)}_{L-1} z^{(i)}_{L-1}) + \vec{b}_{L-1}$. \textbf{Right:} A multi-layer feed-forward neural network for a two class classification problem. Yellow: input layer; Blue: hidden layers; Red: biases; Green: output layer. }
		\label{figure:neuron}
	\end{figure}

	The standard feed-forward neural network (also called multi-layer perceptron) is composed of an input layer, an output layer and a varying number of hidden layers (Figure \ref{figure:neuron}). Each layer consists of several elements called \textit{neurons}. Each neuron's \emph{potential} is the weighted sum of its inputs from the previous layer plus a \emph{bias} $b \in \mathbb{R}$:
	\begin{equation}
	potential(\vec{w}, \vec{x}) = z = (\sum_{i=1}^{n} w^{(i)} x^{(i)} ) + b 
	\end{equation}
	$w^{(i)}$ denotes the $i$-th component of the \emph{weight vector} associated with the $i$-th component of the \emph{input vector} $\vec{x}$.
        The $potential(\vec{w}, \vec{x})$  (also denoted as $z \in \mathbb{R}$) serves as in input to the
      \textit{activation function} $\phi(\cdot)$.  The \emph{activation} of the neuron is given by $a = \phi(z) \in \mathbb{R}$. So, a single neuron in any layer takes the two vectors $\vec{w}$ and $\vec{x}$ as inputs, adds a bias $b$ to it and passes the result through the activation function. Since each layer consists of a varying number of such neurons, the outputs of a single layer can be written as 
	\begin{equation}
	\vec{z} = W \cdot \vec{x} + \vec{b}
	\end{equation}
	where $\cdot$ denotes the standard matrix multiplication, $\vec{z} \in \mathbb{R}^m$, $\vec{x} \in \mathbb{R}^n$, $b \in \mathbb{R}^n$ and $W \in \mathbb{R}^{m \times n}$. The values returned by the last layer, the output layer, are denoted by $\vec{y}$. In the classification setting described here, $\vec{y} \in \mathbb{R}^c$, where c denotes the number of classes. Then a neural network can be represented as: 
	\begin{equation}
	\vec{y} = \phi (x) = \phi_L(... \phi_{2}(\phi_1(\vec{x}))
	\end{equation}
	where $\phi_i$, $i = 1, ..., L$, denotes the activation in the $i$-th layer of the neural network and $\vec{x}$ denotes the input to the first hidden layer; see also Figure~\ref{figure:neuron}. 
        
        Research from \cite{szegedy_2014_intriguing_properties_of_nns} shows that neural networks are vulnerable to adversarial attacks. These are non-randomly perturbed inputs to the network so that the wrong class label is predicted,
        see Figure \ref{figure:adv_example}. Regardless of the nature of the attack all methods for generating adversarial examples try to change the output $\vec{y}$ of a neural network.
        For example, in the setting of image classification, this is done by adding a vector $\vec{r} \in \mathbb{R}^n$ to the input $\vec{x} \in \mathbb{R}^n$ of the neural network, while keeping $||\vec{r}||_p$ as small as possible with respect to a user specified $p$-norm\footnote{$p$-norm: $||\vec{x}||_p = (\sum_{i=1}^{n}||x_i||^p)^{\frac{1}{p}}$}:
	\begin{equation} \label{formula:adver_opt}
	min \; ||\vec{r}||_p \; such \; that \; \vec{y} = f(\vec{x}) \neq f(\vec{x+r}),
	\end{equation}
        where	$\vec{y}$ denotes the true class of the input $\vec{x}$ and $f(\cdot)$ denotes the neural network in general.


        In the following sections, we will illustrate the available verification methods using a toy running example --
        the famous \emph{``and-gate''} Perceptron by~\cite{McP43}:

        \begin{example}[Perceptron]\label{ex:per}
The training set is given by the truth values of the logical connective \emph{and}, see Figure~\ref{figure:toy}.
          \begin{figure}[t]
            \centering
 $\xymatrix@R=0.5pc@C=0pc{
&*\txt{$A$}\ar[rrrr]&&&&*\txt{$w_{A}$}\ar[rrrrd]&&&&&&&&\\
&*\txt{$B$}\ar[rrrr]&&&&*\txt{$w_{B}$}\ar[rrrr]&&&&*+++[o][F]{b_{and}}\ar[rrr]&&&*\txt{$out_{and}$}&\\
} $          \ \ \ \ \ \ \  
            \begin{tabular}{|ll |c |}
\hline
 A & B & A \emph{and} B  \\ \hline
1 & 1 & 1  \\
1 & 0 & 0 \\
0 & 1 & 0 \\
0 & 0 & 0 \\
\hline 
\end{tabular}
\caption{Toy running example. \textbf{Left:} Perceptron computing the logical connective \emph{and}. \textbf{Right:} The ``training data set'' for \emph{and}. }
		\label{figure:toy}
              \end{figure}
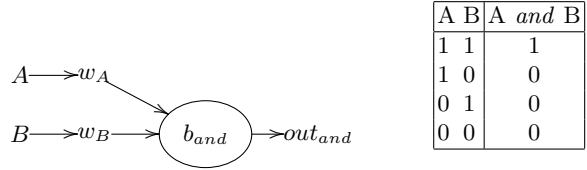

              The Perceptron is trained to  approximate a suitable linear function, e.g. $ out_{and} (A, B) =  b_{and} + w_A \times A + w_B \times B$, given numbers $A$ and $B$ as inputs. E.g. $ -0.9 + 0.5 A + 0.5 B$ would be a suitable solution.
              Sometimes with the Perceptron, we use an activation function
              $S$ given by a threshold function: 
\[
    S(x)= 
\begin{cases}
    1, & \text{if } x\geq 0\\
    0,              & \text{otherwise}
\end{cases}
\]

In the next few sections we assume a verification scenario in which we prove that, subject to certain constraints on values of $A$ and $B$,
the Perceptron will output ``correct'' value for \emph{and}. For example, we may want to prove that it always outputs $1$, when the input values of $A$ and $B$ are greater than $0.7$ (the $0.3$ slack will model the region in which an adversarial manipulation may occur).             
\end{example}


\section{Z3 API in Python and Its Limitations}\label{sec:Py}

Typically, AI students are well-versed in Python.
Thus, the most accessible method of neural verification for them is the method
based on Z3 API in Python. It can give rather quick results, without adding an additional burden of learning a new programming language.
Learning Z3 syntax is a hurdle but not a major challenge, as first-order logic is usually taught in AI courses.
Moreover, Z3 API can be used irrespective of the Python library in which the students implement their networks. 
The Python/Z3 verification workflow will generally comprise of the following steps:

   \begin{itemize}
   \item Implement the neural network in Python. 
     \begin{example}\label{ex:PP}
      We define in Python  the Perceptron given 
      in Figure~\ref{figure:toy}, and use  the \emph{and} training set to train it; see the code snippet in Figure~\ref{figure:PyP}.
      Python computes a suitable Perceptron model with weights
        \begin{center}  \lstinline{ [-0.18375655  0.19388244  0.19471828] }, \end{center} i.e. computing $b_{and} = -0.18375655 $, $w_A = 0.19388244$, and $w_B = 0.19471828$.
       \end{example}
      \item Prove it is robust for a given output class:
        \begin{enumerate}
      \item Define its robustness region: e.g., assume $1$ is the desired output given the input array that contains real values in 
        the region $[0.7 ; 1.5]$; cf. conditions \texttt{cond1} and \texttt{cond2} in Figure~\ref{figure:PyV}.
      \item Define a step function (``the ladder'') to generate a finite number of adversarial examples
        in this region (the finiteness condition is needed to ensure Z3 termination). See e.g. lines 9-13 in Figure~\ref{figure:PyV}.
      \item Prove that the ladder is ``covering'', i.e. guarantees that no adversarial example exists in this region if no adversarial example is found for the ladder. These proofs are
         ``pen and paper''.
      \item  Take the set of adversarial examples  generated by Z3 and run them through the Perceptron.
       If  no misclassification is detected  -- we have proven the Perceptron robust for the given output, region and  ladder.
\end{enumerate}
\end{itemize}

          \begin{figure}[t]
            \centering
\footnotesize{
            \begin{tabular}{l }
\hline
\begin{lstlisting}
class Perceptron(object):
  def __init__(self, eta=0.01, n_iter=50, random_state=1):
        self.eta = eta
        self.n_iter = n_iter
        self.random_state = random_state

  def fit(self, X, y):
  ...
   def predict(self, X):
   return np.where(self.net_input(X) >= 0.0, 1, 0)
   ...
ppn = Perceptron(eta=0.1, n_iter=10)
res = ppn.fit(X,y) 
\end{lstlisting}

              \\
\hline 
\end{tabular}}
\caption{A snippet of Python code implementing a Perceptron.}
		\label{figure:PyP}
              \end{figure}

          \begin{figure}[t]
            \centering
\footnotesize{
            \begin{tabular}{l }
\hline
\begin{lstlisting}
 from z3 import *

X = [ Real("x_%s" % (i+1)) for i in range(2) ]
x1 = X[0]
x2 = X[1]
cond1 = 0.7 <= X[0], X[0] <= 1.5
cond2 = 0.7 <= X[1], X[1] <= 1.5

epsilon = 0.1
ladder = [x*0.5 for x in range(1,20)]

cond3 = Or ([X[0] == l * epsilon for l in ladder])
cond4 = Or ([X[1] == l * epsilon for l in ladder])

s = Solver()
s.add(cond1, cond2, cond3, cond4)

count = 0
while s.check() == sat:
    m = s.model()
    r = [simplify(m.evaluate(X[i])).as_decimal(20) for i in range(2)]
    pred = ppn.predict(list(map(float, r)))
    print("Perceptron predicts: ", pred)
    if pred == 1:
        print("all is ok")
    else:
        print("Counter-example found!")
        break
        s.add(Or(x1 != s.model()[x1], x2 != s.model()[x2])) 
\end{lstlisting}

              \\
\hline 
\end{tabular}}
\caption{A snippet of Python code implementing the \emph{and-gate} Perceptron verification.}
		\label{figure:PyV}
              \end{figure}

              The shown toy example code can be adapted straightforwardly for arbitrary complex neural network implementation.
              Indeed, exercises that follow this simple methodology will be introduced in the academic year 2019-20
              in the standard course of \emph{Data Mining and Machine Learning}  for MSc students with AI specialisms at Heriot-Watt University.
\emph{Simplicity of use is thus the first benefit of choosing Z3 Python API.}              

This methodology has been used in complex extensions, including  e.g.~\cite{HuangKWW17,SinghGPV19}. There, deep convolutional neural networks were verified via Z3 Python API, and featured generation of adversarial examples using more sophisticated Z3 conditions,
that describe in finer detail the functions used in the hidden layers.
\emph{Potential for use in state-of-the-art, industrial scenarios is thus the second benefit of this verification methodology.}  

LAIV already has experience of supervising MSc projects~\cite{K19,LH19} that are devoted to the systematic study of properties of neural networks
considered in~\cite{HuangKWW17}. The \emph{Deep Learning Verification (DLV)}
software that accompanies~\cite{HuangKWW17} is complex enough to provide sufficient challenge for strong
MSc students in their independent projects.
For the same reasons, such projects may not be suitable for an average student who may be easily discouraged by the need to debug and understand the complex Python/Z3 code.  

At the same time, we see several drawbacks, in the long term, for taking Z3-Python API approach
as the main and only methodology for teaching neural network verification:

\begin{itemize}
\item Z3 has no direct access to Python's objects, which is due to a big difference in syntax and semantics of Python and Z3 as
  programming languages.
\end{itemize}  
  \begin{example}
    We return to the Perceptron  defined in Example~\ref{ex:per}.
    Looking at lines 1-13 of Figure~\ref{figure:PyV} we first see the code that defines constraints for Z3 models and does not directly refer 
    to the Perceptron defined in  Figure~\ref{figure:PyP}. Lines 15-19 pass these constraints to Z3 and obtain the model.
    Then lines 21 and 22 show the backwards translation of Z3 models
    into floating-number vectors that are acceptable by Python. The rest of the code in that Figure merely tests the generated  vectors using Python's objects.
\end{example}
Such methodology, irrespective of sophistication of the neural networks and the verification conditions involved,
cannot in principle be extended
  to verification scenarios that state and prove  properties of the implemented neural networks directly. Therefore, the use of Z3 API in Python will always be restricted to first generating some data (subject to constraints)
  via Z3, and then testing that data using the Python's objects.
  Such style of verification, due to the indirect connection between the object we verify and the verification conditions we formulate,
  will itself be prone to errors\footnote{Which will very likely occur rather often if this becomes a  routine methodology used by masses of AI specialists that graduate from the universities.}.  
\begin{itemize}
\item This methodology will inherently suffer from fragility of conversion between symbolic objects that represent real numbers in Z3
  and Python's floating point numbers.
\end{itemize}
  \begin{example}
    Changing the line 21 in  Figure~\ref{figure:PyV} to
    
    \lstinline{r = [simplify(m.evaluate(X[i])).as_decimal(10) for i in range(2)]}

  will cause an error:

   \lstinline{ValueError: could not convert string to float: '0.9500000000?'}
    \end{example}
    Ultimately, if this methodology is used in mass production, all sorts of bugs such as above may become an impediment
    to both soundness and productivity of neural network verification.
\begin{itemize}
  \item Finally, the penultimate item in the methodology described above, the ``pen and paper'' proof of
    adequacy of the chosen region and ladder, will not scale well for
    industrial uses. Manual proofs require an additional production time and are themselves prone to error\footnote{The pen and paper proof is missing in ~\cite{HuangKWW17}.}.
    
\end{itemize}
              
\section{ITP Approach to Neural Network Verification}\label{sec:Coq}

An alternative approach, that would resolve all three shortcomings identified in the previous section, is the ITP approach, as
advocated in e.g.~\cite{BS19,BMF18}.
It amounts to first defining neural networks directly in Coq, and then proving properties about these definitions directly,
thus avoiding programming in Python at the verification stage. Automating translation from Python classes into Coq definitions at the pre-processing stage is possible, and was done in~\cite{BS19}. This translation phase is however not an essential part of the verification process.

          \begin{figure}[t]
            \centering
\footnotesize{
            \begin{tabular}{l }
\hline
        \begin{lstlisting}
Record Neuron := MakeNeuron {
  Output: list nat;
  Weights: list Q;
  Bias: Q;
  Potential: Q;
  Output_Bin: Bin_List Output;
  PosBias: Qlt_bool 0 Bias = true;
  WRange: WeightInRange Weights = true
}.
 \end{lstlisting}
              \\
\hline 
\end{tabular}}
\caption{Coq record type defining a Perceptron.}
		\label{figure:CoP}
              \end{figure}

\begin{example}\label{ex:CoP}
  Figure~\ref{figure:CoP} shows the Perceptron (of Example~\ref{ex:per}) defined in Coq as a record type. We use notation that is very close to~\cite{BMF18}.  
  We note that, in addition to direct declaration of types for outputs, weights,  bias and potential, the type already allows us to
  also assert some additional properties: e.g. that the output must be binary, bias must be positive, and weights must be in range $[-1, 1]$, the latter is defined by a boolean function   \lstinline{WeightInRange}.
  The types    \lstinline{nat} and  \lstinline{Q} are natural and rational numbers respectively. Rational numbers are taken here instead of
  the real numbers, due to complications one would need to face when working with real numbers in Coq (cf. e.g. \cite{Cohen12}).

  The following is an example of a concrete Perceptron (of type  \lstinline{Neuron}) that satisfies the above definition: 
{\small{
  \begin{lstlisting}
Lemma Perceptron : Neuron.
Proof.
  apply (MakeNeuron [1%nat] [2#10; 2#10] (2#10) 1); simpl; auto.
Qed.
   \end{lstlisting}}}
   It has output given by the list $[1]$ (this corresponds to $out_{and}$ of Example~\ref{ex:per}), weights $w_A$ and $w_B$ given by $[\frac{2}{10}, \frac{2}{10}]$, bias $b_{and}$ -- by ${\frac{2}{10}}$, and potential -- by $1$. This roughly corresponds to the Perceptron computed by Python in Example~\ref{ex:PP}, subject to rounding of Python floating numbers to rationals. Note that $b$ must be
   positive according to the Neuron record, the subtraction effect is incorporated via a slightly different threshold function computing $out_{and}$ (this is purely a matter of convenience). 
      
  One can then state and prove lemmas of varying degree of generality (note that the \lstinline{Neuron} type of Figure~\ref{figure:CoP} defines Perceptron's in general, rather than any concrete Perceptron). 
  For example, one may want to prove that, feeding any input of the correct type to a Perceptron, we always get the next output within a permissible range.
  Figure~\ref{fig:fix} shows a function that computes a neuron's potential, and defines the notion of the ``next output''. The Lemma then has a rather simple statement and proof~\cite{BMF18}:

       \begin{figure}[t]
            \centering
\footnotesize{
            \begin{tabular}{l }
\hline
        \begin{lstlisting}
Fixpoint potential (Weights: list Q) (Inputs: list nat): Q :=
  match Weights, Inputs with
  | nil, nil => 0
  | nil, _ => 0
  | _, nil => 0
  | h1::t1, h2::t2 => if (beq_nat h2 0%nat)
                        then (potential t1 t2)
                        else (potential t1 t2) + h1
                        end.

Definition NextPotential (N: Neuron) (Inputs: list nat): Q :=
  if (Qle_bool (Bias N) (Potential N))
      then  (potential (Weights N) Inputs)
      else  (potential (Weights N) Inputs) +  (Potential N).


Definition NextOutput (N: Neuron) (Inputs: list nat) : nat :=
  if (Qle_bool (Bias N) (NextPotential N Inputs))
      then 1%nat
      else 0%nat.
 \end{lstlisting}
              \\
\hline 
\end{tabular}}
\caption{Coq function that computes neuron's potential; formal definition of the notion of the ``next output'', cf.~\cite{BMF18}.}
		\label{fig:fix}
              \end{figure}

  {\small{
   \begin{lstlisting}
Lemma NextOutput_Bin_List: forall (N: Neuron) (Inputs: list nat),
   Bin_List(Output N) -> Bin_List (NextOutput N Inputs::Output N).
Proof.
  intros. simpl. split.
  - unfold NextOutput. destruct (Qle_bool (Bias N) (NextPotential N Inputs)).
    + simpl. reflexivity.
    + simpl. reflexivity.  - apply H.
Qed.
   \end{lstlisting}}}
 
\end{example}

To summarise the benefits of ITP approach to neural network verification:

\begin{itemize}
\item it makes definitions of neural networks, and all associated functions, fully transparent not just for the software engineer
  who works with the code, but also for the type checker, that certifies soundness of the code;
\item it ensures that when proving properties of neural networks, we directly refer to their implementation in the same language as opposed to the indirect ``testing'' style of verification that we had to resort to in Python and Z3.
\item we no longer need any ``pen and paper proofs'' in the verification cycle, as Coq's language is rich enough to state lemmas of arbitrary generality;
\item finally, we no longer have any problems with intra-language translation and type mismatch that was pertinent to the Z3 --  Python API approach. 

\end{itemize}  

The Coq approach to neural network verification also has drawbacks, especially if we consider taking it as the main
verification tool for AI and Robotics students. Firstly, there is a serious overhead arising from the need to learn Coq from scratch.
For a student whose first and main programming language is Python, Coq has
an unusual syntax, and requires a non-trivial degree of understanding of the associated logic and
type theory for successful proof completion. If one day Coq and similar provers are introduced as compulsory element of AI degrees,
this would become feasible. As it stands, immediate introduction of Coq-based neural network verification in classes or projects is not realistic.

Among other pitfalls of this approach are:

\begin{itemize}
\item real numbers are not as accessible in Coq as floating numbers are in Python. The rational numbers as used in all the running examples of this section are not a feasible substitute for real numbers in the long run, as many machine learning algorithms rely on real-numbered differentiable functions.
  Even a more sophisticated approach of~\cite{BS19} shies away from real numbers, an thus has to claim only working with ``quantised'' neural networks.
\item Although Coq is in principle  a programming language rather than just a prover, actual computations with its functions are problematic.
  \begin{example}\label{ex:prob}
    Let us try to use the Perceptron definition given in Example~\ref{ex:CoP}, and define its next outputs given the input vector $[1,1]$:
  {\small{  \begin{lstlisting}
Definition Pp : nat :=
  NextOutput Perceptron [1%nat; 1%nat].
     \end{lstlisting}}}
     Now trying to actually compute the number (using the command \lstinline{Compute (Pp).}) gives a few hundreds lines of an output evaluating
     Coq functions
     (a snippet is  shown in Figure~\ref{fig:out}), --
     something a Python programmer will be baffled with!
   \end{example}

 \item An instance of the same problem of function evaluation arises in trying to prove adversarial robustness of individual neural networks (in the style of Python-Z3 approaches). Such verification will inevitably depend on evaluating functions and computing outputs of the networks.      

   \begin{example}\label{ex:fp}
     Continuing with the same Coq file, and attempting to prove the correctness of the output of the Perceptron defined in Example~\ref{ex:CoP}
     given the input $[1, 1]$ will result in the following lemma:
{\small{\begin{lstlisting}
Lemma sound: forall x y : nat,  x = 1%nat ->  y = 1%nat ->  (NextOutput Perceptron [x ; y]) = 1%nat.
\end{lstlisting}}}
   Its direct proof by simplification is unfortunately infeasible, for the reasons explained in Example~\ref{ex:prob}.  
     \end{example}

 \end{itemize}

       \begin{figure}[t]
            \centering
\footnotesize{
            \begin{tabular}{p{12cm} }
\hline
        \begin{lstlisting}
    = if
        match
          match
            match (let (Qnum, _) := let (Output, Weights, Bias, _, _, _, _) := Perceptron in Bias in Qnum) with
            | 0%Z => 0%Z
            | Z.pos x =>
                Z.pos
                  ((fix Ffix (x0 x1 : positive) {struct x0} : positive :=
                      match x0 with
                      | (x2~1)%positive =>
                          (fix Ffix0 (x3 x4 : positive) {struct x3} : positive :=
                             match x3 with
                             | (x5~1)%positive =>
                             ...
 \end{lstlisting}
              \\
\hline 
\end{tabular}}
\caption{A snippet of Coq's output computing function Pp.}
		\label{fig:out}
              \end{figure}

              We thus have to conclude that there are still insufficient grounds for arguing Coq's suitability as a mainstream language of verification for masses of AI and machine learning specialists.

\section{$F^*$: A Hybrid Approach to Neural Net Verification}\label{sec:fstar}

In this section we investigate whether new hybrid languages, such as Microsoft's $F^*$~\cite{MartinezADGHHNP19}, bear promise to resolve
major drawbacks of the two approaches we surveyed so far.
$F^*$ is a general-purpose functional programming language with effects, that combines
the automation of an SMT solver Z3
   with dependent types.
After verification, F* programs can be extracted to efficient OCaml, $F\#$, C, WASM, or ASM code. 

Much of $F^*$'s verification of neural networks would follow the ITP (type-driven) methodology as explained in Section~\ref{sec:Coq}.
For example, the lemma from Example~\ref{ex:CoP} could be stated and proven in $F^*$.
However, unlike Coq, $F^*$ offers a crucial benefit for machine learning verification projects: it has a user-friendly implementation of real numbers,
that connects to Z3's library for  \emph{linear  real arithmetic}; same library on which Z3-Python approach of Section~\ref{sec:Py} relies upon! 

\begin{example}
  Figure~\ref{fig:FP} shows  $F^*$ definition of a \lstinline{neuron} type. It looks very similar to Coq code in Figure~\ref{figure:CoP}.
  But, since $F^*$ now allows us to use real numbers easily,
 \lstinline{neuron} definition takes full advantage of this: see e.g. line 8 of
Figure~\ref{fig:FP} which defines a constraint for the real-valued bias.
\end{example}

          \begin{figure}[t]
            \centering
\footnotesize{
            \begin{tabular}{l }
\hline
        \begin{lstlisting}
 noeq type neuron =  
 | MakeNeuron : 
  output: list nat
  -> weights: list real
  -> bias: real
  -> potential: real
  -> output_Bin: bin_list output
  ->  posbias: 0.0R <=. bias
  ->   wRange: weightinrange weights
->  neuron
 \end{lstlisting}
              \\
\hline 
\end{tabular}}
\caption{$F^*$ record type defining a Perceptron.}
		\label{fig:FP}
              \end{figure}

              The integration of Z3 with a dependently-typed language brings many advantages for neural network verification.
              To start with, direct implementation of real numbers allows a more precise encoding of neural networks.
\begin{example}\label{ex:FSP}
  Our Perceptron as computed by Python in Example~\ref{ex:PP} can now have a direct encoding in $F^*$:
{\small{\begin{lstlisting}
val perceptron : neuron
 let perceptron = MakeNeuron [1] [0.194R ; 0.195R] 0.184R 1.0R  
\end{lstlisting}}}
Compare this with Example~\ref{ex:CoP}, in which we used Coq's rational numbers to approximate floating point numbers of Python.
            \end{example}

Proofs of properties that were impossible in Coq  due to the Coq's implementation of
function evaluation,
now come fully automated
via Z3's theory of linear real arithmetic.

            \begin{example}\label{ex:final}
              We return to the simple soundness property which we failed to prove in Coq (see Example~\ref{ex:fp}.)
              We can now generalise and prove that property. Firstly, we can now formulate it in direct correspondence
              with our original Example~\ref{ex:per}, where we set up our robustness region to be within
              $0.3$ distance of the desired input value $1$. This gives rise to the following rather natural robustness statement for the \lstinline{perceptron} defined in Example~\ref{ex:FSP}:
            {\small{  \begin{lstlisting}
 let add_id_l = assert (forall m n. ( (m >=. 0.7R) /\ (n >=. 0.7R)) ==> (nextoutput perceptron [m ; n]) == 1)
            \end{lstlisting}}}
 Proving this assertion takes no effort, as it is performed by $F^*$'s Z3 API fully automatically.
\end{example}

Importantly, the ability to prove such assertions fully addresses the concerns we raised in Section~\ref{sec:Py} about Python-Z3 verification projects.
In Python, one of the main drawbacks was our inability to directly reason about Python's objects that implemented  neural networks.
This had consequences such as introducing the  ``indirect'' testing style of verification, that was not just prone to all sorts of human and interfacing errors, but also required manual proofs of the soundness properties for the chosen region and ladder.  
As Example~\ref{ex:final} shows, $F^*$'s compiler can give full translation of the properties of $F^*$'s neural network definitions  into Z3 solver. 
Thus, the assertion of  Example~\ref{ex:final}  refers to the object
\lstinline{perceptron} defined in Example~\ref{ex:FSP} directly, reducing the chance of verification errors.
It moreover states a property concerning the
real-valued input signals without any need to define any finite subsets, regions or ladders.
This is again possible thanks to the direct communication with the theory of linear real arithmetic in Z3.
Such direct proofs of properties of the neural network implementation was impossible in Python.

\section{Conclusions and Future Directions}\label{sec:concl}

We have outlined two major verification paths currently available to AI and Robotics students: one is based on the idea
of automated theorem proving integrated into Python; another -- based on an ITP such as Coq.
Rather than focusing on the  well-known advantages and disadvantages of automated and interactive theorem proving in general, we focused
on the verification methodologies these languages dictate when applied to neural networks. In particular, we showed that
automated proving fits better with the idea of verifying adversarial robustness of individual networks, whereas interactive proving is largely unusable for this purpose, yet is much better suited for verification of more general properties of neural networks (defined more generally as types).

We outlined surprising or notable features of these two verification approaches.  Z3 API in Python gives the most accessible verification framework for AI students,
as it requires little effort for those already fluent with Python, and can be easily embedded into
the daily workflow of Python programmers, irrespective of the Python neural network library they use.
However, due to
opacity of Python's objects for Z3, Python-Z3 verification methodology is indirect,
bound to its strong testing flavour and inherently relies on some pen-and-paper proofs for its soundness.
We showed that this particular limitation is easily resolved by defining neural networks directly in Coq, but this approach fails to cater properly for real numbers indispensable in machine learning and has other limitations related to function evaluation. We showed that this limitation, in turn,  can be resolved in a language $F^*$ that is in many ways
similar to Coq, but offers more efficient reasoning about real valued functions. Comparing convenience and power of Z3 API in Python and $F^*$, the latter has the benefit of direct translation of $F^*$ objects into $Z3$ theories, thus allowing to state and prove soundness properties of implemented neural networks directly. On that basis, we see a promise in such languages for teaching  neural network verification.

\bibliographystyle{splncs04}
\bibliography{laiv}

\end{document}